\documentclass[10pt,twocolumn,letterpaper]{article}

\usepackage{cvpr}
\usepackage{times}
\usepackage{epsfig}
\usepackage{graphicx}
\usepackage{amsmath}
\usepackage{amssymb}
\usepackage{paralist}
\usepackage{color}
\usepackage{float}
\usepackage{comment}
\usepackage{multirow}
\usepackage{caption} 
\usepackage{enumitem}
\usepackage[pagebackref=true,breaklinks=true,letterpaper=true,colorlinks,bookmarks=false]{hyperref}

\cvprfinalcopy % *** Uncomment this line for the final submission

\def\cvprPaperID{769} % *** Enter the CVPR Paper ID here

\def\ie{\emph{i.e}\onedot}

\def\etal{\emph{et al}\onedot}

\ifcvprfinal\fi
\begin{document}

%%%%%%%%% TITLE
% \title{Social GAN: Generative Adversarial Networks for Trajectory Prediction}
%\title{Social GAN: Human Trajectory Prediction \\ with Generative Adversarial Networks}
\title{Social GAN: Socially Acceptable Trajectories \\ with Generative Adversarial Networks}
%\title{Socially-accepted Generative Adversarial Network}

\author{
	Agrim Gupta\textsuperscript{1} \hspace{3mm}
    Justin Johnson\textsuperscript{1} \hspace{3mm}
    Li Fei-Fei\textsuperscript{1} \hspace{3mm}
    Silvio Savarese\textsuperscript{1} \hspace{3mm}
    Alexandre Alahi\textsuperscript{1,2} \\*
%     {\tt\small agrim@stanford.edu} \hspace{2mm}
%     {\tt\small jcjohns@cs.stanford.edu} \hspace{2mm}
%     {\tt\small alexandre.alahi@epfl.ch} \hspace{2mm}
%     {\tt\small feifeili@cs.stanford.edu} \\*
     Stanford University\textsuperscript{1} \hspace{5mm}
    \'Ecole Polytechnique F\'ed\'erate de Lausanne\textsuperscript{2} \\*
%     {\tt\small \{jcjohns,alahi,feifeili\}@cs.stanford.edu}
}

\maketitle
%\thispagestyle{empty}

%%%%%%%%% ABSTRACT
\begin{abstract}
% Understanding and predicting human motion behavior is essential for navigating human-centric environments with autonomous moving platforms such as self-driving vehicles or socially-aware robots. Recent data-driven methods suffer from effectively modeling multimodal human-human interactions, \ie, given the same observation, future predictions can be different. In this work, we propose to tackle this challenge with a Generative Adversarial Network (GAN) based encoder-decoder architecture that captures the multimodal distribution of human behavior. The goal of our generator is to output multiple socially-accepted paths. We introduce a new variety loss and a pooling mechanism that outperform previous works in terms of computation complexity and accuracy. We perform thorough analysis of our model in terms of accuracy, variety of prediction, speed and collision avoidance on multiple publicly available datasets. 
Understanding human motion behavior is critical for autonomous moving platforms (like self-driving cars and social robots) if they are to navigate human-centric environments. This is challenging because human motion is inherently multimodal: given a history of human motion paths, there are many socially plausible ways that people could move in the future. We tackle this problem by combining tools from sequence prediction and generative adversarial networks: a recurrent sequence-to-sequence model observes motion histories and predicts future behavior, using a novel pooling mechanism to aggregate information across people. We predict socially plausible futures by training adversarially against a recurrent discriminator, and encourage diverse predictions with a novel variety loss. Through experiments on several  datasets we demonstrate that our approach outperforms prior work in terms of accuracy, variety, collision avoidance, and computational complexity.
\end{abstract}
\vspace{-2mm}

\section{Introduction} \label{sec:intro}
\begin{figure}[t]
\centering
\includegraphics[width=\linewidth]{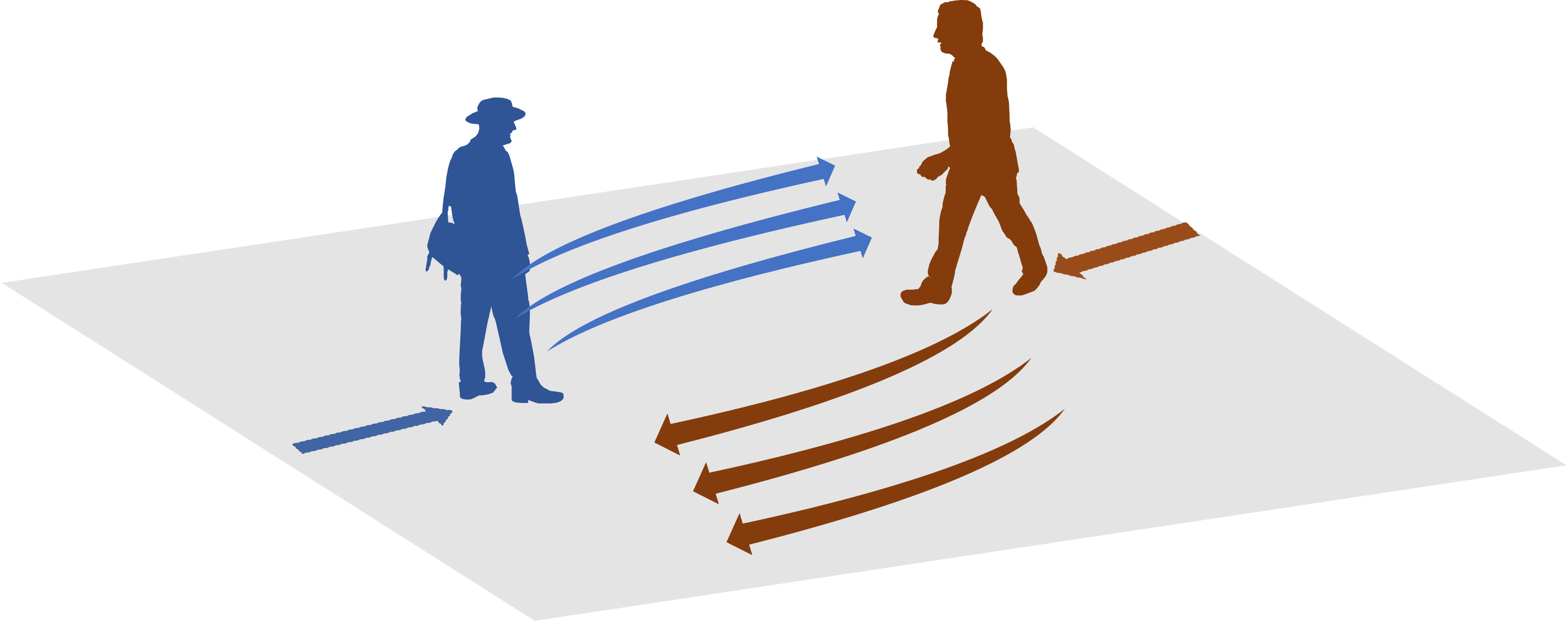}
   \caption{
   		Illustration of a scenario where two pedestrians want to avoid each other. There are many possible ways that they can avoid a potential collision. We present a method that given the same observed past, predicts multiple socially acceptable outputs in crowded scenes. 
   }
\label{fig:pull}
\end{figure}

%\textcolor{red}{Paper as of now is written assuming we have only one pooling mechanism. The system figure shows three. Will add that if I get good results with neighbor pooling.}
Predicting the motion behavior of pedestrians is essential for autonomous moving platforms like self-driving cars or social robots that will share the same ecosystem as humans. 
Humans can effectively negotiate complex social interactions, and these machines ought to be able to do the same. One concrete and important task to this end is the following: given observed motion trajectories of pedestrians (coordinates for the past \textit{e.g.} 3.2 seconds), predict \textit{all} possible future trajectories (Figure~\ref{fig:pull}).
%They need to have the same ability as humans to effectively negotiate complex social interactions. An important task for autonomous agents is therefore to solve the following problem: given the observed motion trajectories of pedestrians (x-y coordinates within the past \textit{e.g}, 3.2 seconds), output \textit{all} possible future trajectories (\textit{e.g.}, 4.8 seconds) (Figure \ref{fig:pull}). 

Forecasting the behavior of humans is challenging due to the inherent properties of human motion in crowded scenes:
% socially-aware mobility. Human motion trajectories in crowded scenes are:
\begin{enumerate}[leftmargin=*]
  \item \textbf{Interpersonal}. Each person's motion depends on the people around them. Humans have the innate ability to read the behavior of others when navigating crowds. Jointly modeling these dependencies is a challenge.
  \item \textbf{Socially Acceptable}. Some trajectories are physically possible but socially unacceptable. Pedestrians are governed by social norms like yielding right-of-way or respecting personal space. Formalizing them is not trivial.
  \item \textbf{Multimodal}. Given a partial history, there is no single correct future prediction. Multiple trajectories are plausible and socially-acceptable. 
\end{enumerate}
%They are rational agents, \ie they do the right thing as defined by Russel and Norvig in \cite{russell2009artificial}. 
%This ability is essential for autonomous agents that will share the same ecosystem as humans. Self-driving vehicles or social robots will operate along side humans. They need to have the same ability as humans to effectively negotiate complex social interactions and provide a sense of familiarity.
%------------------------------------------------------------------------
%------------------------------------------------------------------------
Pioneering work in trajectory prediction has tackled some of the above challenges. The interpersonal aspect has been exhaustively addressed by traditional methods based on hand-crafted features \cite{antonini2006discrete,helbing1995social,tay2008modelling,yamaguchi2011you}. Social acceptability has been recently revisited with data-driven techniques based on Recurrent Neural Networks (RNNs) \cite{alahi2016social,lee2017desire,fernando2017soft+,bartoli2017context}. Finally, the multimodal aspect of the problem has been studied in the context of route choices given a static scene (\textit{e.g.}, which streets to take at an intersection \cite{lee2017desire,kitani2012activity}). Robicquet \etal \cite{robicquet2016learning} have shown that pedestrians have multiple navigation styles in crowded scenes given a mild or aggressive style of navigation. Therefore, the forecasting task entails outputting different possible outcomes. 

While existing methods have made great progress in addressing specific challenges, they suffer from two limitations. First, they model a local neighborhood around each person when making the prediction. Hence, they do not have the capacity to model interactions between all people in a scene in a computationally efficient fashion. Second, they tend to learn the ``average behavior'' because of the commonly used loss function that minimizes the euclidean distance between the ground truth and forecasted outputs. In contrast, we aim in learning multiple ``good behaviors", \textit{i.e.}, multiple socially acceptable trajectories.

To address the limitations of previous works, we propose to leverage the recent progress in generative models. Generative Adversarial Networks (GANs) have been recently developed to overcome the difficulties in approximating intractable probabilistic computation and behavioral inference \cite{goodfellow2014generative}. While they have been used to produce photo-realistic signals such as images \cite{odena2016conditional}, we propose to use them to generate multiple socially-acceptable trajectories given an observed past. One network (the generator) generates candidates and the other (the discriminator) evaluates them. The adversarial loss enables our forecasting model to go beyond the limitation of L2 loss and potentially learn the distribution of ``good behaviors" that can fool the discriminator. In our work, these behaviors are referred to as \emph{socially-accepted} motion trajectories in crowded scenes. 

Our proposed GAN is a RNN Encoder-Decoder generator and a RNN based encoder discriminator with the following two novelties: (i) we introduce a variety loss which encourages the generative network of our GAN to spread its distribution and cover the space of possible paths while being consistent with the observed inputs. (ii) We propose a new pooling mechanism that learns a ``global'' pooling vector which encodes the subtle cues for all people involved in a scene. We refer to our model as ``Social GAN''.
Through experiments on several publicly available real-world crowd datasets, we show state-of-the-art accuracy, speed and demonstrate that our model has the capacity to generate a variety of socially-acceptable trajectories.
\section{Related Work}

Research in forecasting human behavior can be grouped as learning to predict human-space interactions or human-human interactions. The former learns scene-specific motion patterns \cite{ballan2016knowledge,coscia2016point,hu2007semantic,kim2011gaussian,kitani2012activity,morris2011trajectory,zhou2011random}. The latter models the dynamic content of the scenes, \ie how pedestrians interact with each other. The focus of our work is the latter: learning to predict human-human interactions. We discuss existing work on this topic as well as relevant work in RNN for sequence prediction and Generative models.

\textbf{Human-Human Interaction.}  Human behavior has been studied from a crowd perspective in \textit{macroscopic models} or from a individual perspective in \emph{microscopic models} (the focus of our work). One example of microscopic model is the Social Forces by Helbing and Molnar \cite{helbing1995social} which models pedestrian behavior with attractive forces guiding them towards their goal and repulsive forces encouraging collision avoidance. Over the past decades, this method has been often revisited \cite{choi2012unified,choi2014understanding,leal2014learning,leal2011everybody,luber2010people,mehran2009abnormal,pellegrini2010improving,yamaguchi2011you}. Tools popular in economics have also been used such as the Discrete Choice framework by Antonini \textit{et. al.} \cite{antonini2006discrete}. Treuille \textit{et. al.} \cite{treuille2006continuum} use continuum dynamics, and Wang \textit{et. al.} \cite{wang2008gaussian}, Tay \textit{et. al.} \cite{tay2008modelling} use Gaussian processes. Such functions have also been used to study stationary groups \cite{parksocial15,yi2015understanding}. However, all these methods use hand crafted energy potentials based on relative distances and specific rules. In contrast, over the past two years, data-driven methods based on RNNs have been used to outperform the above traditional ones.

\textbf{RNNs for Sequence Prediction.} Recurrent Neural Networks are a rich class of dynamic models which extend feedforward networks for sequence generation in diverse domains like speech recognition \cite{chorowski2014end,chung2015recurrent,graves2014towards}, machine translation \cite{chung2015recurrent} and image captioning \cite{karpathy2014deep, vinyals2015show, xu2015show, shu2017cern}. However, they lack high-level and spatio-temporal structure \cite{liu2016spatio}. Several attempts have been made to use multiple networks to capture complex interactions \cite{alahi2016social, du2015hierarchical, srivastava2015unsupervised}. Alahi \etal\cite{alahi2016social} use a social pooling layer that models nearby pedestrians. In the rest of this paper, we show that using a Multi-Layer Perceptron (MLP) followed by max pooling is computationally more efficient and works as well or better than the social pooling method from \cite{alahi2016social}. Lee \textit{et al.} \cite{lee2017desire} introduce a RNN Encoder-Decoder framework which uses variational auto-encoder (VAE) for trajectory prediction. However, they did not model human-human interactions in crowded scenes. 
%------------------------------------------------------------------------
\begin{figure*}[t!]
\centering
\includegraphics[width=0.93\linewidth]{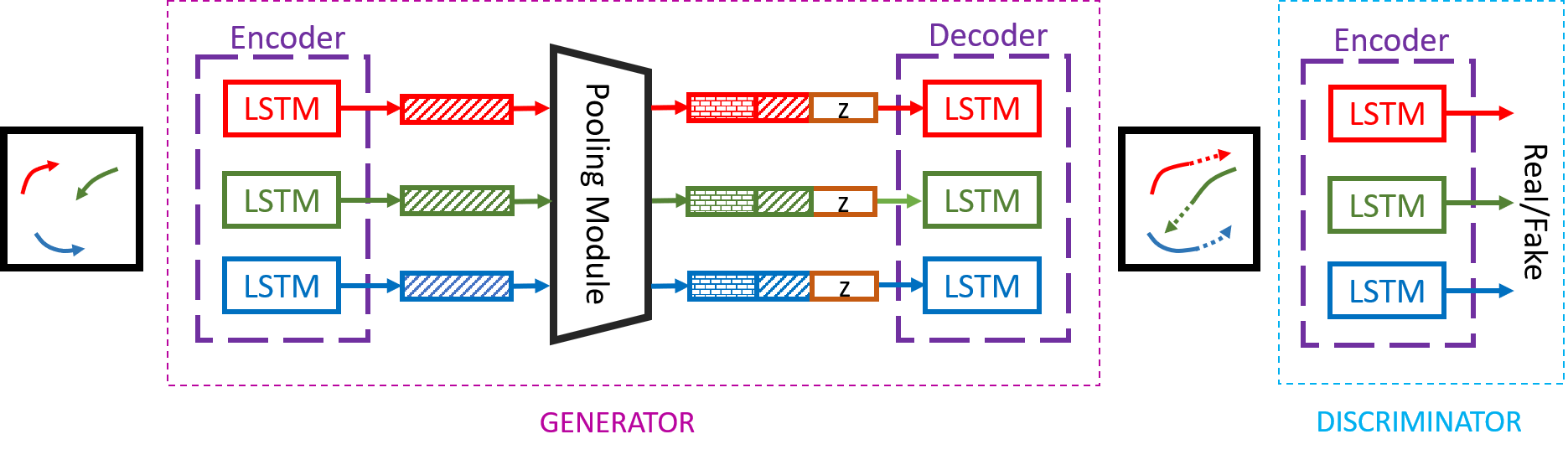}
\vspace{-4mm}
   \caption{
     System overview. Our model consists of three key components: Generator (G), Pooling Module, and Discriminator (D). G takes as input past trajectories $X_i$ and encodes the history of the person $i$ as $H_i^t$. The pooling module takes as input all $H_i^{t_{obs}}$ and outputs a pooled vector $P_i$ for each person. The decoder generates the future trajectory conditioned on $H_i^{t_{obs}}$ and $P_i$. D takes as input $T_{real}$ or $T_{fake}$ and classifies them as socially acceptable or not (see Figure \ref{fig:pool} for PM).
   }
\label{fig:system}
\end{figure*}
%------------------------------------------------------------------------

\textbf{Generative Modeling.} Generative models like variational autoencoders \cite{kingma2013auto} are trained by maximizing the lower bound of training data likelihood. Goodfellow \etal\cite{goodfellow2014generative} propose an alternative approach, Generative Adversarial Networks (GANs), where the training procedure is a minimax game between a generative model and a discriminative model; this overcomes the difficulty of approximating intractable probabilistic computations. Generative models have shown promising results in tasks like super-resolution \cite{ledig2016photo}, image to image translation \cite{isola2016image}, and image synthesis \cite{gregor2015draw, odena2016conditional, zhang2016stackgan} which have multiple possible outputs for a given input. However, their application in sequence generation problems like natural language processing has lagged since sampling from these generated outputs to feed to the discriminator is a non-differentiable operation. 

\section{Method}
Humans possess an intuitive ability to navigate crowds taking into account the people around them. We plan our paths keeping in mind our goal and also simultaneously taking into account the motion of surrounding people like their direction of motion, velocity, etc. However, often in such situations multiple possible options exist. We need  models which not only can understand these complex human interactions but can also capture the variety of options. Current approaches have focused on predicting the average future trajectory which minimizes the $L2$ distance from the ground truth future trajectory whereas we want to predict multiple ``good'' trajectories. In this section, we first present our GAN based encoder-decoder architecture to address this issue, we then describe our novel pooling layer which models human-human interactions and finally we introduce our variety loss which encourages the network to produce multiple diverse future trajectories for the same observed sequence. 

\subsection{Problem Definition}
Our goal is to \textbf{jointly} reason and predict the future trajectories of \textbf{all} the agents involved in a scene. We assume that we receive as input all the trajectories for people in a scene as $\mathbf{X} = {X_1, X_2,..., X_n}$ and predict the future trajectories $\mathbf{\hat{Y}} = {\hat{Y_1}, \hat{Y_2},..., \hat{Y_n}}$ of all the people \textbf{simultaneously}. The input trajectory of a person $i$ is defined as $X_i = {(x_i^t, y_i^t)}$ from time steps $t = 1, ..., t_{obs}$ and the future trajectory (ground truth) can be defined similarly as ${Y_i} = {({x}_i^t, {y}_i^t)}$ from time steps $t= t_{obs} + 1, ... , t_{pred}$. We denote predictions as $\hat{Y_i}$. 

\subsection{Generative Adversarial Networks}
A Generative Adversarial Network (GAN) consists of two neural networks trained in opposition to each other \cite{goodfellow2014generative}. The two adversarially trained models are: a generative model $G$ that captures the data distribution, and a discriminative model $D$ that estimates the probability that a sample came from the training data rather than $G$. The generator $G$ takes a latent variable $z$ as input, and outputs sample $G(z)$. The discriminator $D$ takes a sample $x$ as input and outputs $D(x)$ which represents the probability that it is real. The training procedure is similar to a two-player min-max game with the following objective function:
%------------------------------------------------------------------------
\begin{equation} 
\begin{aligned}\label{eq:gan}
& \min_G\max_D V(G, D) = \\
& \mathbb{E}_{x \sim p_{data}(x)}[\log D(x)] + \mathbb{E}_{z \sim p_{(z)}}[\log(1 - D(G(z)))].
\end{aligned}
\end{equation}
%------------------------------------------------------------------------
GANs can used for conditional models by providing both the generator and discriminator with additional input $c$,
% if both the generator and discriminator are conditioned on some extra information $c$, 
yielding $G(z, c)$ and $D(x, c)$ \cite{gauthier2014conditional, mirza2014conditional}.
\subsection{Socially-Aware GAN}
As discussed in Section \ref{sec:intro} trajectory prediction is a multi-modal problem. Generative models can be used with time-series data to simulate possible futures. We leverage this insight in designing SGAN which addresses the multi-modality of the problem using GANs (see Figure \ref{fig:system}). Our model consists of three key components: Generator (G), Pooling Module (PM) and Discriminator (D). G is based on encoder-decoder framework where we link the hidden states of encoder and decoder via PM. G takes as input $X_i$ and outputs predicted trajectory $\hat{Y}_i$. D inputs the entire sequence comprising both input trajectory $X_i$ and future prediction $\hat{Y}_i$ (or $Y_i$) and classifies them as ``real/fake''.

\textbf{Generator.} We first embed the location of each person using a single layer MLP to get a fixed length vector $e_i^t$. These embeddings are used as input to the LSTM cell of the encoder at time $t$ introducing the following recurrence:
%------------------------------------------------------------------------
\begin{equation} 
\begin{aligned}\label{eq:encoder}
e_i^t &= \phi(x_i^t, y_i^t; W_{ee}) \\
h_{ei}^t &= LSTM(h_{ei}^{t-1}, e_i^t; W_{encoder})
\end{aligned}
\end{equation}
%------------------------------------------------------------------------
where $\phi(\cdot)$ is an embedding function with ReLU non-linearity, $W_{ee}$ is the embedding weight. The LSTM weights ($W_{encoder}$) are shared between all people in a scene. 

Na\"ive use of one LSTM per person fails to capture interaction between people. Encoder learns the state of a person and stores their history of motion. However, as shown by Alahi \etal\cite{alahi2016social} we need a compact representation which combines information from different encoders to effectively reason about social interactions. In our method, we model human-human interaction via a Pooling Module (PM). After $t_{obs}$ we pool hidden states of all the people present in the scene to get a pooled tensor $P_i$ for each person. Traditionally, GANs take as input noise and generate samples. Our goal is to produce future scenarios which are consistent with the past. To achieve this we condition the generation of output trajectories by initializing the hidden state of the decoder as:
%------------------------------------------------------------------------
\begin{equation} 
\begin{aligned}\label{eq:encoder}
c_i^t &= \gamma(P_i, h_{ei}^t; W_{c}) \\
h_{di}^t &= [c_i^t, z]
\end{aligned}
\end{equation}
%------------------------------------------------------------------------
where $\gamma(\cdot)$ is a multi-layer perceptron (MLP) with ReLU non-linearity and $W_{c}$ is the embedding weight. We deviate from prior work in two important ways regarding trajectory prediction:
\begin{itemize}[noitemsep,topsep=0pt,parsep=0pt,partopsep=0pt]
\item Prior work \cite{alahi2016social} uses the hidden state to predict parameters of a bivariate Gaussian distribution. However, this introduces difficulty in the training process as backpropagation through sampling process in non-differentiable. We avoid this by directly predicting the coordinates $(\hat{x}_i^t, \hat{y}_i^t)$. 
\item ``Social'' context is generally provided as input to the LSTM cell \cite{alahi2016social,lee2017desire} . Instead we provide the pooled context only once as input to the decoder. This also provides us with the ability to choose to pool at specific time steps and results in \textbf{16x} speed increase as compared to S-LSTM \cite{alahi2016social} (see Table \ref{table:speed}). 
\end{itemize}
After initializing the decoder states as described above we can obtain predictions as follows:
%------------------------------------------------------------------------
\begin{equation} 
\begin{aligned}\label{eq:encoder}
e_i^t &= \phi(x_i^{t-1}, y_i^{t-1}; W_{ed}) \\
P_i &= PM(h_{d1}^{t-1},...,h_{dn}^t) \\
h_{di}^t &= LSTM(\gamma(P_i, h_{di}^{t-1}), e_i^t; W_{decoder}) \\
(\hat{x}_i^t, \hat{y}_i^t) &= \gamma(h_{di}^t)
\end{aligned}
\end{equation}
%------------------------------------------------------------------------
where $\phi(\cdot)$ is an embedding function with ReLU non-linearity with $W_{ed}$ as the embedding weights. The LSTM weights are denoted by $W_{decoder}$ and $\gamma$ is an MLP.

\textbf{Discriminator}. The discriminator consists of a separate encoder. Specifically, it takes as input $T_{real} = [X_i, Y_i]$ or $T_{fake} = [X_i, \hat{Y}_i]$ and classifies them as real/fake. We apply a MLP on the encoder's last hidden state to obtain a classification score. The discriminator will ideally learn subtle social interaction rules and classify trajectories which are not socially acceptable as ``fake''. \par

\textbf{Losses}. In addition to adversarial loss, we also apply $L2$ loss on the predicted trajectory which measures how far the generated samples are from the actual ground truth.
\subsection{Pooling Module}
%------------------------------------------------------------------------
\begin{figure}[t]
\centering
\includegraphics[width=\linewidth]{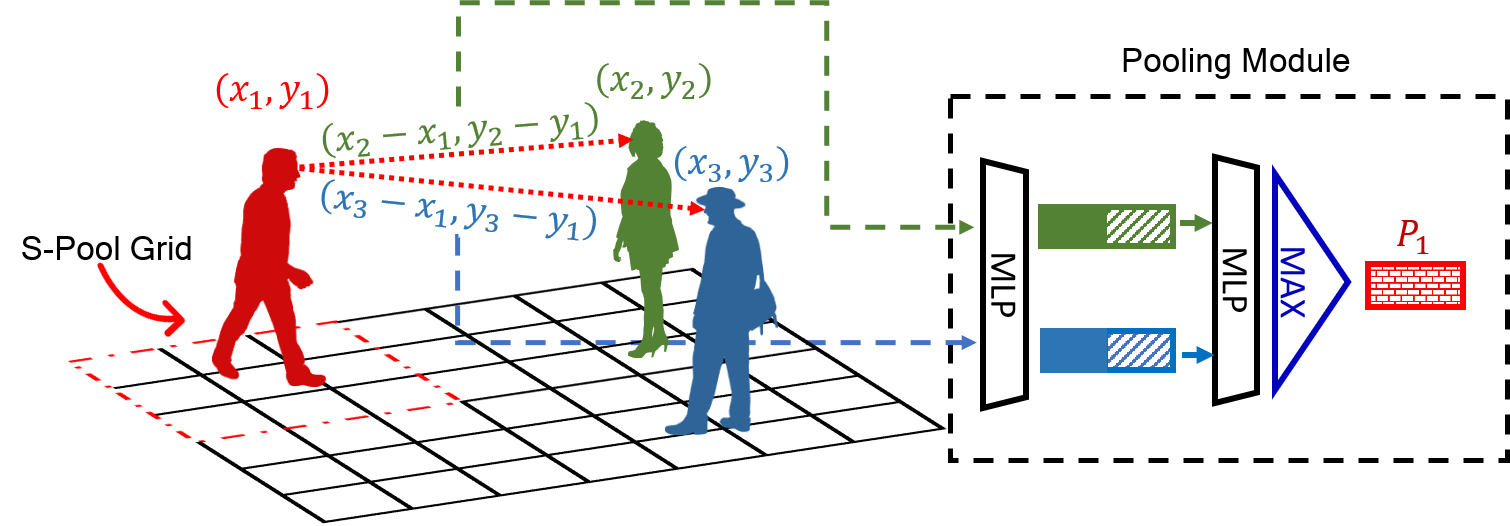}
\caption{
	Comparison between our pooling mechanism (red dotted arrows) and Social Pooling~\cite{alahi2016social} (red dashed grid) for the red person. Our method computes relative positions between the red and all other people; these positions are concatenated with each person's hidden state, processed independently by an MLP, then pooled  elementwise to compute red person's pooling vector $P_1$. Social pooling only considers people inside the grid, and cannot model interactions between all pairs of people.}
   	%We show pooling mechanism for red person. We obtain an embedding of relative position of each person in the scene with respect to red person. This spatial information combined with the associated hidden state is passed through a MLP to obtain the pooling vector $P_1$. Neighborhood pooling mechanism like S-Pooling would ignore the people outside the grid and consequently fail to model complex social interactions.}
\label{fig:pool}
\end{figure}
%------------------------------------------------------------------------
In order to jointly reason across multiple people we need a mechanism to share information across LSTMs. However, there are several challenges which a method should address:

\begin{itemize}[noitemsep,topsep=0pt,parsep=0pt,partopsep=0pt]
\item Variable and (potentially) large number of people in a scene. We need a compact representation which combines information from all the people. 
\item  Scattered Human-Human Interaction. Local information is not always sufficient. Far-away pedestrians might impact each others. Hence, the network needs to model global configuration.  
\end{itemize}

Social Pooling \cite{alahi2016social} addresses the first issue by proposing a grid based pooling scheme. However, this hand-crafted solution is slow and fails to capture global context. Qi \etal\cite{qi2016pointnet} show that above properties can be achieved by applying a learned symmetric function on transformed elements of the input set of points. As shown in Figure \ref{fig:system} this can be achieved by passing the input coordinates through a MLP followed by a symmetric function (we use Max-Pooling). The pooled vector $P_i$ needs to summarize all the information a person needs to make a decision. Since, we use relative coordinates for translation invariance we augment the input to the pooling module with relative position of each person with respect to person $i$.
\begin{table*}[ht!]
  \centering
   \setlength{\tabcolsep}{2.4mm}
   \renewcommand{\arraystretch}{1.1}
  \begin{tabular}{c|c|c|c|c|c|c|c|c}
  \multirow{2}{*}{\textbf{Metric}} & \multirow{2}{*}{\textbf{Dataset}} & \multirow{2}{*}{\textbf{Linear}} & \multirow{2}{*}{\textbf{LSTM}} & \multirow{2}{*}{\textbf{S-LSTM}} & \multicolumn{4}{|c}{\textbf{SGAN (Ours)}} \\
  & & & & \cite{alahi2016social} & 1V-1 & 1V-20 & 20V-20 & 20VP-20 \\ 
  \hline
  \hline
  \multirow{5}{*}{ADE} & \textbf{ETH} & 0.84 / 1.33 & 0.70 / 1.09 & 0.73 / 1.09 &  0.79 / 1.13  & 0.75 / 1.03 &  0.61 / \textbf{0.81} &  \textbf{0.60} / 0.87 \\
  & \textbf{HOTEL} &  \textbf{0.35} / \textbf{0.39} & 0.55 / 0.86 & 0.49 / 0.79 & 0.71 / 1.01 & 0.63 / 0.90 &  0.48 / 0.72 & 0.52 / 0.67 \\
  & \textbf{UNIV} & 0.56 / 0.82 & 0.36 / 0.61 & 0.41 / 0.67 & 0.37 / 0.60 & 0.36 / 0.58 &  \textbf{0.36} / \textbf{0.60} & 0.44 / 0.76 \\
  & \textbf{ZARA1} & 0.41 / 0.62 & 0.25 / 0.41  &  0.27 / 0.47  & 0.25 / 0.42 & 0.23 / 0.38 & \textbf{0.21} / \textbf{0.34} & 0.22 / 0.35 \\ 
  & \textbf{ZARA2} & 0.53 / 0.77 & 0.31 / 0.52 & 0.33 / 0.56 &  0.32 / 0.52 & 0.29 / 0.47 & \textbf{0.27} / 0.42 & 0.29 / \textbf{0.42} \\
  \hline
  \textbf{AVG} &  & 0.54 / 0.79 & 0.43 / 0.70 & 0.45 / 0.72 & 0.49 / 0.74 & 0.45 / 0.67 & \textbf{0.39} / \textbf{0.58} & 0.41 / 0.61 \\
  \hline 
  \hline
  \multirow{5}{*}{FDE} & \textbf{ETH} & 1.60 / 2.94 & 1.45 / 2.41 & 1.48 / 2.35 & 1.61 / 2.21 & 1.52 / 2.02 & 1.22 / \textbf{1.52} & \textbf{1.19} / 1.62 \\
  & \textbf{HOTEL} & \textbf{0.60} / \textbf{0.72} & 1.17 / 1.91 & 1.01 / 1.76 & 1.44 / 2.18 & 1.32 / 1.97 & 0.95 / 1.61 & 1.02 / 1.37 \\
  & \textbf{UNIV} & 1.01 / 1.59 & 0.77 / 1.31 & 0.84 / 1.40 & 0.75 / 1.28 & \textbf{0.73} / \textbf{1.22} & 0.75 / 1.26 & 0.84 / 1.52 \\
  & \textbf{ZARA1} & 0.74 / 1.21 & 0.53 / 0.88 & 0.56 / 1.00 & 0.53 / 0.91 & 0.48 / 0.84 & \textbf{0.42} / 0.69 & 0.43 / \textbf{0.68} \\
  & \textbf{ZARA2} & 0.95 / 1.48 & 0.65 / 1.11 & 0.70 / 1.17 & 0.66 / 1.11 & 0.61 / 1.01 & \textbf{0.54} / \textbf{0.84} & 0.58 / 0.84 \\
  \hline 
  \textbf{AVG} & & 0.98 / 1.59 & 0.91 / 1.52 & 0.91 / 1.54 & 1.00 / 1.54 & 0.93 / 1.41 & \textbf{0.78} / \textbf{1.18} & 0.81 / 1.21 \\ 
  \hline \hline
  \end{tabular}
  \caption{Quantitative results of all methods across datasets. We report two error metrics Average Displacement Error (ADE) and Final Displacement Error (FDE) for $t_{pred} = 8$ and $t_{pred} = 12$ (8 / 12) in meters. Our method consistently outperforms state-of-the-art S-LSTM method and is especially good for long term predictions (lower is better).}
  \label{table:quant}
\end{table*}
%------------------------------------------------------------------------
\subsection{Encouraging Diverse Sample Generation}
Trajectory prediction is challenging as given limited past history a model has to reason about multiple possible outcomes. The method described so far produces good predictions, but these predictions try to produce the ``average'' prediction in cases where there can be multiple outputs. Further, we found that outputs were not very sensitive to changes in noise and produced very similar predictions. \par 
We propose a variety loss function that encourages the network to produce diverse samples. For each scene we generate $k$ possible output predictions by randomly sampling $z$ from $\mathcal{N}(0, 1)$ and choosing the ``best'' prediction in $L2$ sense as our prediction. 
%------------------------------------------------------------------------
\begin{equation} 
\begin{aligned}\label{eq:gan}
\mathcal{L}_{variety} = \min_k \Vert Y_i - \hat{Y}_i^{(k)} \Vert_2,
\end{aligned}
\end{equation}
%------------------------------------------------------------------------
where $k$ is a hyperparameter. \par 
By considering only the best trajectory, this loss encourages the network to hedge its bets and cover the space of outputs that conform to the past trajectory. The loss is structurally akin to Minimum over N (MoN) loss \cite{fan2016point} but to the best of our knowledge this has not been used in the context of GANs to encourage diversity of generated samples. 

\subsection{Implementation Details}
We use LSTM as the RNN in our model for both decoder and encoder. The dimensions of the hidden state for encoder is 16 and decoder is 32. We embed the input coordinates as $16$ dimensional vectors. We iteratively train the Generator and Discriminator with a batch size of 64 for 200 epochs using Adam \cite{kingma2014adam} with an initial learning rate of 0.001. %We used PyTorch for our implementation. 
%The model is trained on a single Tesla K80.

\section{Experiments}
In this section, we evaluate our method on two publicly available datasets: ETH \cite{pellegrini2010improving} and UCY \cite{leal2014learning}. These datasets consist of real world human trajectories with rich human-human interaction scenarios. We convert all the data to real world coordinates and interpolate to obtain values at every $0.4$ seconds. In total there are $5$ sets of data (ETH - 2, UCY-3) with $4$ different scenes which  consists of $1536$ pedestrians in crowded settings with challenging scenarios like group behavior, people crossing each other, collision avoidance and groups forming and dispersing. \par
\textbf{Evaluation Metrics}. Similar to prior work \cite{alahi2016social,lee2017desire} we use two error metrics:
\begin{enumerate}[leftmargin=*]
  \setlength\itemsep{0mm}
\item \emph{Average Displacement Error (ADE)}: Average $L2$ distance between ground truth and our prediction over all predicted time steps.
\item \emph{Final Displacement Error (FDE)}: The distance between the predicted final destination and the true final destination at end of the prediction period $T_{pred}$.
\end{enumerate} \par
\textbf{Baselines}: We compare against the following baselines:
\begin{enumerate}[leftmargin=*]
  \setlength\itemsep{0mm}
\item \emph{Linear}: A linear regressor that estimates linear parameters by minimizing the least square error. 
\item \emph{LSTM}: A simple LSTM with no pooling mechanism.
\item \emph{S-LSTM}: The method proposed by Alahi \etal\cite{alahi2016social}. Each person is modeled via an LSTM with the hidden states being pooled at each time step using the social pooling layer. 
\end{enumerate}
We also do an ablation study of our model with different control settings. We refer our full method in the section as \emph{SGAN-kVP-N} where $kV$ signifies if the model was trained using variety loss ($k = 1$ essentially means no variety loss) and $P$ signifies usage of our proposed pooling module. At test time we sample multiple times from the model and chose the best prediction in $L2$ sense for quantitative evaluation. $N$ refers to the number of  time we sample from our model during test time. \par
%------------------------------------------------------------------------
\begin{figure*}[t]
\centering
\includegraphics[width=1\linewidth]{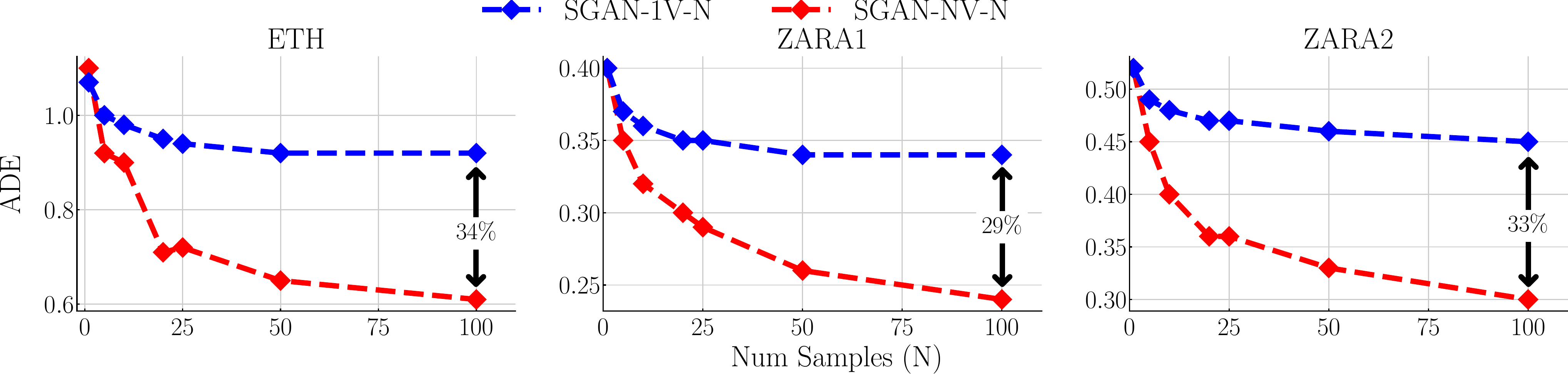}
	\caption{Effect of variety loss. For SGAN-1V-N we train a single model, drawing one sample for each sequence during training and $N$ samples during testing. For SGAN-NV-N we train several models with our variety loss, using $N$ samples during both training and testing. Training with the variety loss significantly improves accuracy.}
   % \caption{We analyze the effect of variety loss on model accuracy. In case of SGAN-1V-N, simply drawing more samples doesn't lead to any performance benefit. However, as we vary k performance quickly improves till $k=25$ and then plateaus. Please note that each red point is a separate model. On average models with $k=100$ perform $32\%$ better than SGAN-1V-100 counterpart across datasets.}
\label{fig:variety}
\end{figure*}
%------------------------------------------------------------------------
\textbf{Evaluation Methodology}. We follow similar evaluation methodology as \cite{alahi2016social}. We use leave-one-out approach, train on $4$ sets and test on the remaining set. We observe the trajectory for $8$ times steps ($3.2$ seconds) and show prediction results for $8$ (3.2 seconds) and $12$ (4.8 seconds) time steps.
\subsection{Quantitative Evaluation}
We compare our method on two metrics ADE and FDE against different baselines in Table \ref{table:quant}. As expected Linear model is only capable of modeling straight paths and does especially bad in case of longer predictions ($t_{pred} = 12$). Both LSTM and S-LSTM perform much better than the linear baseline as they can model more complex trajectories. However, in our experiments S-LSTM does not outperform LSTM. We tried our best to reproduce the results of the paper. \cite{alahi2016social} trained the model on synthetic dataset and then fine-tuned on real datasets. We don't use synthetic data to train any of our models which could potentially lead to worse performance. \par 
SGAN-1V-1 performs worse than LSTM as each predicted sample can be any of the multiple possible future trajectories. The conditional output generated by the model represents one of many plausible future predictions which might be different from ground truth prediction. When we consider multiple samples our model outperforms the baseline methods confirming the multi-modal nature of the problem. GANs face mode collapse problem, where the generator resorts to generating a handful of samples which are assigned high probability by the discriminator. We found that samples generated by SGAN-1V-1 didn't capture all possible scenarios. However, SGAN-20V-20 significantly outperforms all other models as the variety loss encourages the network to produce diverse samples. Although our full model with proposed pooling layer performs slightly worse we show in the next section that pooling layer helps the model predict more ``socially'' plausible paths. \par
\textbf{Speed.} Speed is crucial for a method to be used in a real-world setting like autonomous vehicles where you need accurate predictions about pedestrian behavior. We compare our method with two baselines LSTM and S-LSTM. A simple LSTM performs the fastest but can't avoid collisions or make accurate multi-modal predictions. Our method is \textbf{16x} faster than S-LSTM (see Table \ref{table:speed}). Speed improvement is because we don't do pooling at each time step. Also, unlike S-LSTM which requires computing a occupancy grid for each pedestrian our pooling mechanism is a simple MLP followed by max pooling. In real-world applications our model can quickly generate $20$ samples in the same time it takes S-LSTM to make $1$ prediction. \par
\textbf{Evaluating Effect of Diversity.} One might wonder what will happen if we simply draw more samples from our model without the variety loss? We compare the performance of SGAN-1V-N with SGAN-NV-N. As a reminder SGAN-NV-N refers to a model trained with variety loss with $k = N$ and drawing $N$ samples during testing. As shown in Figure \ref{fig:variety} across all datasets simply drawing more samples from the model trained without variety loss does not lead to better accuracy. Instead, we see a significant performance increase as we increase $k$ with models on average performing $33\%$ better with $k=100$ .
%------------------------------------------------------------------------
\begin{table}
  \centering
   \setlength{\tabcolsep}{2.2mm}
   \renewcommand{\arraystretch}{1.1}
  \begin{tabular}{c|c|c|c|c}
   & \textbf{LSTM} & \textbf{S-LSTM} & \textbf{SGAN} & \textbf{SGAN-P} \\
  \hline
  \hline
  8 & 0.02 & 1.79 & 0.04 & 0.12 \\
  12 & 0.03 & 2.61 & 0.05 & 0.15 \\
  \hline
  Speed-Up & 82x & 1x & \textbf{49x} & \textbf{16x} \\ \hline \hline
  \end{tabular}
  \caption{Speed (in seconds) comparison with S-LSTM. We get 16x speedup as compared to S-LSTM allowing us to draw 16 samples in the same time S-LSTM makes a single prediction. Unlike S-LSTM we don't perform pooling at each time step resulting in significant speed bump without suffering on accuracy. All methods are benchmarked on Tesla P100 GPU}
  \label{table:speed}
\end{table}
%------------------------------------------------------------------------
\subsection{Qualitative Evaluation}
\def\ratio{0.96}
\begin{figure*}[ht]
  \centering
    \includegraphics[trim={0px 0px 0px 0px},width=\ratio\textwidth,clip=true]{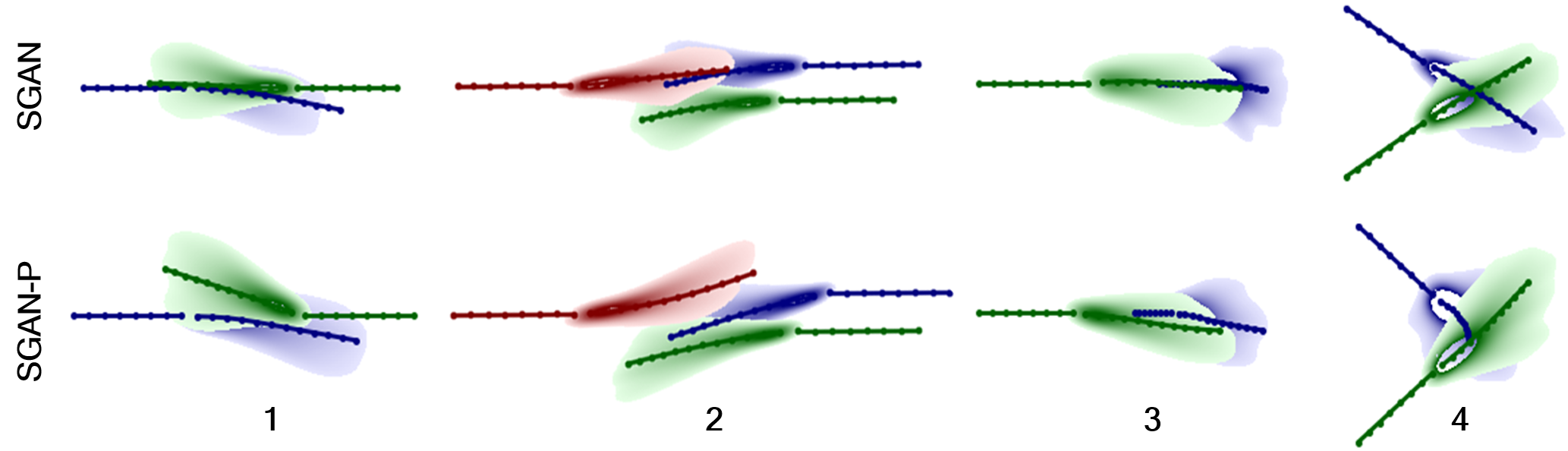}
    \caption{Comparison between our model without pooling (SGAN, top) and with pooling (SGAN-P, bottom) in four collision avoidance scenarios: two people meeting (1), one person meeting a group (2), one person behind another (3), and two people meeting at an angle (4). For each example we draw 300 samples from the model and visualize their density and mean. Due to pooling, SGAN-P predicts socially acceptable trajectories which avoid collisions.}
  % \caption{Comparison between SGAN-P vs SGAN. We compare how both methods perform in $4$ common collision avoidance scenarios. We show approximate distribution of predicted trajectories and average predicted trajectory. SGAN-P is able to predict subtle changes in direction and speed to achieve collision avoidance in a socially acceptable manner.}
  \label{fig:qual-pool}
\end{figure*}
%------------------------------------------------------------------------
In multi-agent (people) scenarios, it is imperative to model how actions of one person can influence the actions of other people. Traditional approaches for activity forecasting and human trajectory prediction have focused on hand crafted energy potentials modeling attractive and repulsive forces to model these complex interactions. We use a purely data driven approach which models human-human interaction via a novel pooling mechanism. Humans walking in the presence of other people plan their path taking into account their personal space, perceived potential for collision, final destination and their own past motion. In this section, we first evaluate the effect of the pooling layer and then analyze the predictions made by our network in three common social interaction scenarios. Even though our model makes \textbf{joint} predictions for \textbf{all} people in a scene we show predictions for a subset for simplicity. We refer to each person in the scene by the first letter of the color in the figure (e.g., Person B (Black), Person R (Red) and so on). Also for simplicity we refer SGAN-20VP-20 as SGAN-P and SGAN-20V-20 as SGAN. 
\subsubsection{Pooling Vs No-Pooling}
On quantitative metrics both methods perform similarly with SGAN slightly outperforming SGAN-P (see Table \ref{table:quant}. However, qualitatively we find that pooling enforces a global coherency and conformity to social norms. We compare how SGAN and SGAN-P perform in four common social interaction scenarios (see Figure \ref{fig:qual-pool}). We would like to highlight that even though these scenarios were created synthetically, we used models trained on real world data. Moreover, these scenarios were created to evaluate the models and nothing in our design makes these scenarios particularly easy or hard. For each setup we draw $300$ samples and plot an approximate distribution of trajectories along with average trajectory prediction. \par
Scenario 1 and 2 depict the collision avoidance capacity of our model by changing direction. In the case of two people heading in the same direction pooling enables the model to predict a socially accepted way of yielding the right of way towards the right. However, SGAN prediction leads to a collision. Similarly, unlike SGAN, SGAN-P is able to model group behavior and predict avoidance while preserving the notion of couple walking together (Scenario 2). \par
Humans also tend to vary pace to avoid collisions. Scenario 3 is depicts a person G walking behind person B albeit faster. If they both continue to maintain their pace and direction they would collide. Our model predicts person G overtaking from the right. SGAN fails to predict a socially acceptable path.  In Scenario 4, we notice that the model predicts person B slowing down and yielding for person G. 
\begin{comment}
\emph{Case 1}: Consider what happens when a person approaches a group of $2$. SGAN-P recognizes the potential collision and shifts the paths towards right for both the approaching parties. In contrast, SGAN doesn't alter the course and predicts that people would continue in the same manner. Pooling helps the network learn social norms like giving way to the right which is exhibited by the group as well as the person G. Ideally, in case of unbiased data one could expect a bimodal density plot. However, we observe unimodal avoidance behavior with model. \par
\emph{Case 2}: Again we consider the case of two people heading towards each other. In case of no pooling the trajectories are almost straight depicted as string of high density yellow blobs. Please note that high density in the middle indicates collision as both trajectories pass through the same point. SGAN-P is able to detect the presence of other person and alter course accordingly albeit slightly. Unlike reactive models like Social Forces our model produces more smooth and natural looking trajectories.
\end{comment}
\subsubsection{Pooling in Action}
%------------------------------------------------------------------------
\def\ratio{0.233}
\begin{figure*}[ht!]
  \centering
  \includegraphics[trim={0px 0px 0px 0px},width=0.9\textwidth,clip=true]{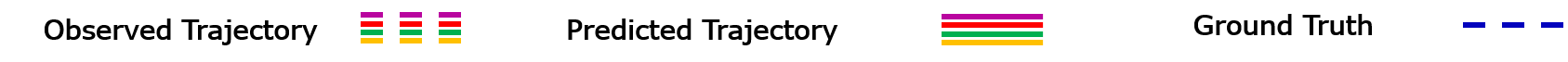} \\
  \hspace{-5mm}
  \raisebox{9mm}{
  	\begin{minipage}{7mm}
  	\centering
  	\rotatebox{90}{\parbox{1.6cm}{\centering \textbf{People} \\ \textbf{Merging}}}
  	\end{minipage}
  }
  \includegraphics[trim={0px 50px 20px 100px},width=\ratio\textwidth,clip=true]{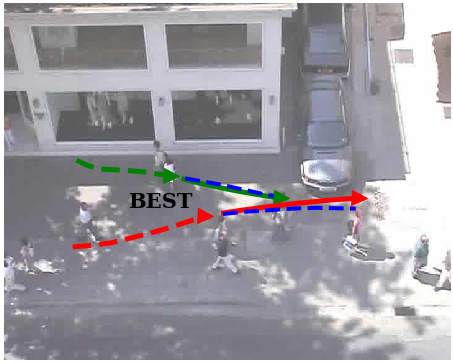}
  \includegraphics[trim={0px 50px 20px 100px},width=\ratio\textwidth,clip=true]{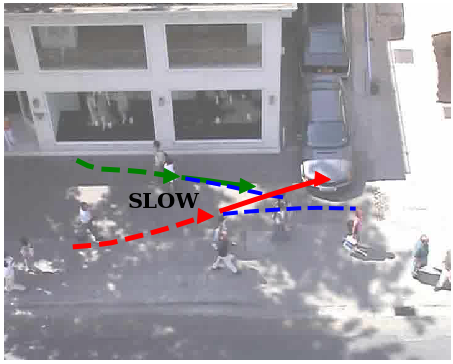}
  \includegraphics[trim={0px 50px 20px 100px},width=\ratio\textwidth,clip=true]{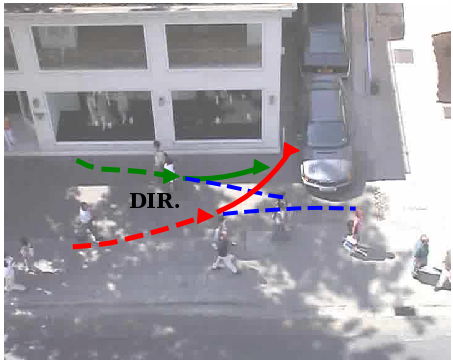}
  \includegraphics[trim={0px 50px 20px 100px},width=\ratio\textwidth,clip=true]{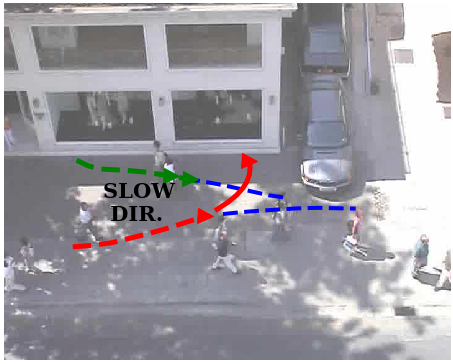}\\
  \vspace{1mm}
  \hspace{-5mm}
  \raisebox{9mm}{
  	\begin{minipage}{7mm}
  	\centering
  	\rotatebox{90}{\parbox{1.6cm}{\centering \textbf{Group} \\ \textbf{Avoiding}}}
  	\end{minipage}
  }
  \includegraphics[trim={0px 50px 20px 100px},width=\ratio\textwidth,clip=true]{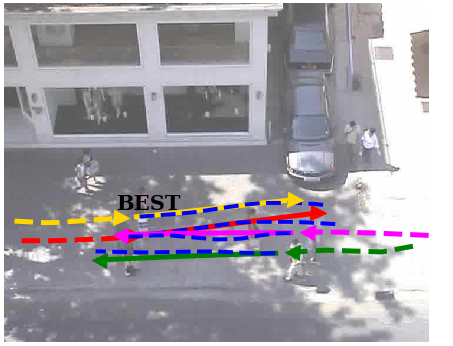}
  \includegraphics[trim={0px 50px 20px 100px},width=\ratio\textwidth,clip=true]{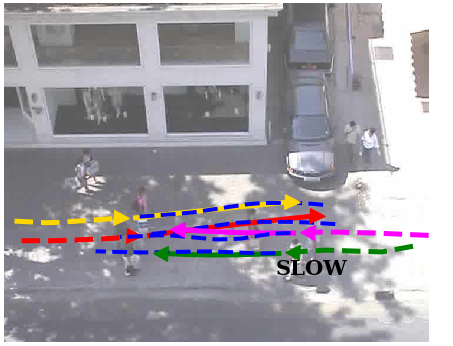}
  \includegraphics[trim={0px 50px 20px 100px},width=\ratio\textwidth,clip=true]{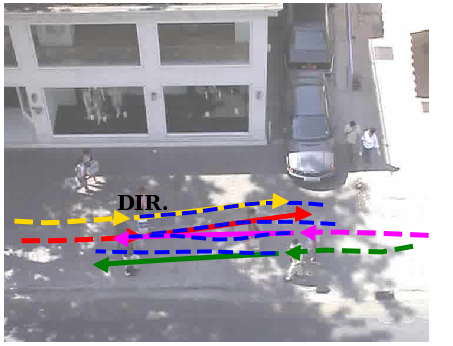}
  \includegraphics[trim={0px 50px 20px 100px},width=\ratio\textwidth,clip=true]{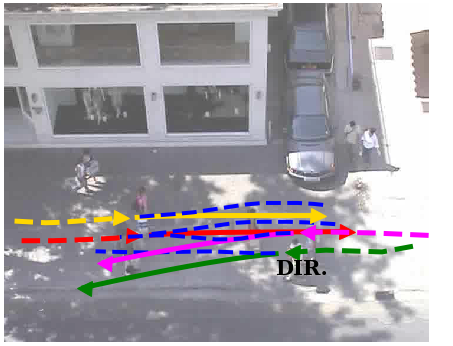}\\
  \vspace{1mm}
  \hspace{-5mm}
  \raisebox{9mm}{
  	\begin{minipage}{7mm}
  	\centering
  	\rotatebox{90}{\parbox{1.6cm}{\centering \textbf{Person} \\ \textbf{Following}}}
  	\end{minipage}
  }
  \includegraphics[trim={0px 50px 20px 100px},width=\ratio\textwidth,clip=true]{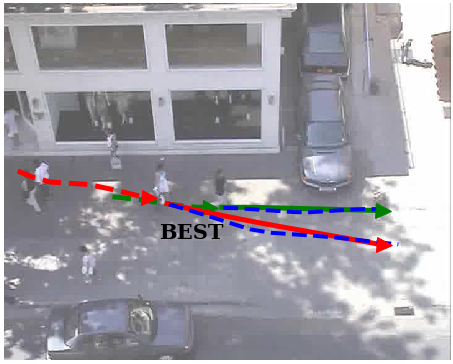}
  \includegraphics[trim={0px 50px 20px 100px},width=\ratio\textwidth,clip=true]{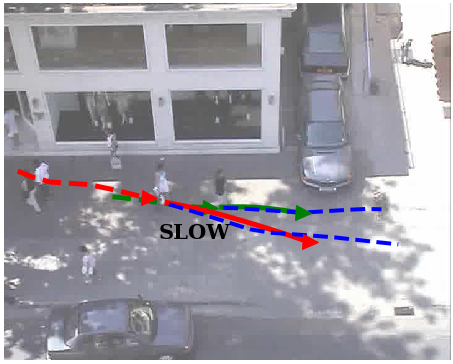}
  \includegraphics[trim={0px 50px 20px 100px},width=\ratio\textwidth,clip=true]{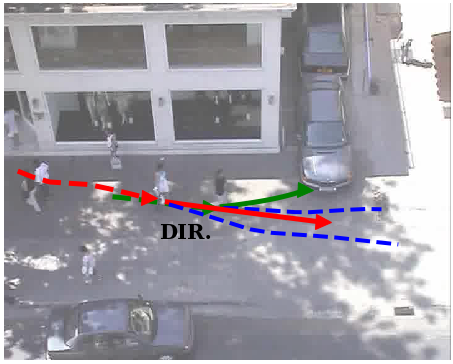}
  \includegraphics[trim={0px 50px 20px 100px},width=\ratio\textwidth,clip=true]{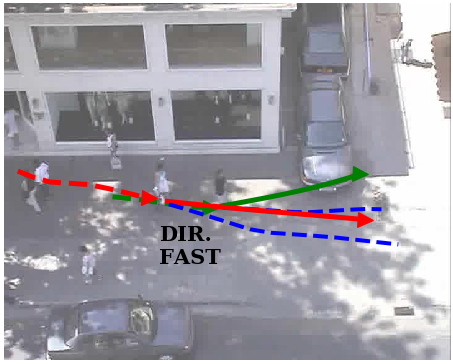}\\
  \hspace{6mm}
  \begin{minipage}{\ratio\textwidth}
    \centering \textbf{(1)}
  \end{minipage}
    \begin{minipage}{\ratio\textwidth}
	\centering \textbf{(2)}
  \end{minipage}
  \begin{minipage}{\ratio\textwidth}
    \centering \textbf{(3)}
  \end{minipage} 
  \begin{minipage}{\ratio\textwidth}
    \centering \textbf{(4)}
  \end{minipage} 
  \caption{
  	Examples of diverse predictions from our model. Each row shows a different set of observed trajectories; columns
    show four different samples from our model for each scenario which demonstrate different types of socially acceptable behavior. BEST is the sample closest to the ground-truth; in SLOW and FAST samples, people change speed
    to avoid collision; in DIR samples people change direction to avoid each other. Our model learns these different avoidance strategies in a data-driven manner, and jointly predicts globally consistent and socially acceptable trajectories for all people in the scene. We also show some failure cases in supplementary material. 
  }
  %\caption{Each row is a sequence from the dataset; for each sequence our model makes many predictions and each column shows four different samples from our model exhibiting different types of socially acceptable future behavior which we label as: 1. BEST: Closest match to ground truth prediction. 2. SLOW/FAST: Change speed to avoid collision. 3. DIR.: Change direction or course to avoid collision. Our method is able to learn these subtle social norms in a purely data-driven manner and jointly predicts trajectories for all people in a scene in a globally consistent and socially acceptable manner.}
  \label{fig:qual-real}
\end{figure*}
%------------------------------------------------------------------------
We consider three real-scenarios where people have to alter their course to avoid collision (see Figure \ref{fig:qual-real}). \par
\textbf{People Merging.} (Row 1) In hallways or in roads it is common for people coming from different directions to merge and walk towards a common destination. People use various ways to avoid colliding while continuing towards their destination. For instance a person might slow down, alter their course slightly or use a combination of both depending on the context and behavior of other surrounding people. Our model is able predict variation in both speed and direction of a person to effectively navigate a situation. For instance model predicts that either person B slows down (col 2) or  both person B and R change direction to avoid collision. The last prediction (col 4) is particularly interesting as the model predicts a sudden turn for person R but also predicts that person B significantly slows down in response; thus making a globally consistent prediction.\par 
\textbf{Group Avoiding}. (Row 2) People avoiding each other when moving in opposite direction is another common scenario. This can manifest in various forms like a person avoiding a couple, a couple avoiding a couple etc. To make correct predictions in such cases a person needs to plan ahead and look beyond it's immediate neighborhood. Our model is able to recognize that the people are moving in groups and model group behavior. The model predicts change of direction for either groups as a way of avoiding collision (col 3, 4). In contrast to Figure \ref{fig:qual-pool} even though the convention might be to give way to the right in this particular situation that would lead to a collision. Hence, our models makes prediction where couples give way towards the left. \par
\textbf{Person Following}. (Row 3) Another common scenario is when a person is walking behind someone. One might want to either maintain pace or maybe overtake the person in front. We would like to draw attention to a subtle difference between this situation and its real-life counterpart. In reality a person's decision making ability is restricted by their field of view. In contrast, our model has access to ground truth positions of all the people involved in the scene at the time of pooling. This manifests in some interesting cases (see col 3). The model understands that person R is behind person B and is moving faster. Consequently, it predicts that person B gives way by changing their direction and person R maintains their direction and speed. The model is also able to predict overtaking (matching the ground truth). 
\subsection{Structure in Latent Space}
\begin{figure}[t]
\centering
\includegraphics[width=0.94\linewidth]{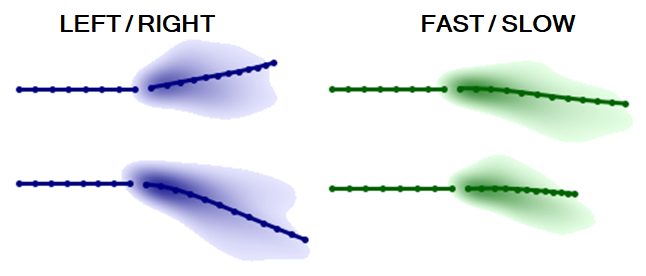}
\caption{Latent Space Exploration. Certain directions in the latent manifold are associated with direction (left) and speed (right). Observing the same past but varying the input $z$ along different directions causes the model to predict trajectories going either right/left or fast/slow on average.}
\label{fig:latent}
\end{figure}

In this experiment we attempt to understand the landscape of the latent space $z$. Walking on the manifold that is learnt can give us insights about how the model is able to generate diverse samples. Ideally, one can expect that the network imposes some structure in the latent space. We found that certain directions in the latent space were associated with direction and speed (Figure \ref{fig:latent}). 
%------------------------------------------------------------------------
%------------------------------------------------------------------------
\section{Conclusion}
In this work we tackle the problem of modeling human-human interaction and jointly predicting trajectories for all people in a scene. We propose a novel GAN based encoder-decoder framework for trajectory prediction capturing the multi-modality of the future prediction problem. We also propose a novel pooling mechanism enabling the network to learn social norms in a purely data-driven approach. To encourage diversity among predicted samples we propose a simple variety loss which coupled with the pooling layer encourages the network to produce globally coherent, socially compliant diverse samples. We show the efficacy of our method on several complicated real-life scenarios where social norms must be followed.
%------------------------------------------------------------------------
\section{Acknowledgment}
We thank Jayanth Koushik and De-An Huang for their helpful comments and suggestions.
{\small
\bibliographystyle{ieee}
\bibliography{egbib}
}

\end{document}